\documentclass[pdflatex, iicol]{sn-jnl}

\usepackage{graphicx}%
\usepackage{multirow}%
\usepackage{amsmath,amssymb,amsfonts}%
\usepackage{amsthm}%
\usepackage{mathrsfs}%
\usepackage[title]{appendix}%
\usepackage{xcolor}%
\usepackage{textcomp}%
\usepackage{manyfoot}%
\usepackage{booktabs}%
\usepackage{algorithm}%
\usepackage{algorithmicx}%
\usepackage{algpseudocode}%
\usepackage{listings}%
\newcommand{\p}[1]{\smallskip \noindent \textbf{{#1}.}}%
\newcommand{\eq}[1]{Equation~(\ref{eq:#1})}%
\newcommand{\fig}[1]{Figure~\ref{fig:#1}}%

\theoremstyle{thmstyleone}%
\theoremstyle{thmstyletwo}%

\theoremstyle{thmstylethree}%

\begin{document}

\title[Article Title]{Should Collaborative Robots be Transparent?}


\author*[1]{\fnm{Shahabedin} \sur{Sagheb}}\email{shahab@vt.edu}

\author[1]{\fnm{Soham} \sur{Gandhi}}\email{sgandhi25@vt.edu}

\author[1]{\fnm{Dylan P.} \sur{Losey}}\email{losey@vt.edu}

\affil*[1]{\orgdiv{Department of Mechanical Engineering}, \orgname{Virginia Tech}, \orgaddress{\street{635 Prices Fork Road}, \city{Blacksburg}, \postcode{24061}, \state{Virginia}, \country{USA}}}


\abstract{We often assume that robots which collaborate with humans should behave in ways that are transparent (e.g., legible, explainable). These transparent robots intentionally choose actions that convey their internal state to nearby humans: for instance, a transparent robot might exaggerate its trajectory to indicate its goal. But while transparent behavior seems beneficial for human-robot interaction, is it actually \textit{optimal}? In this paper we consider collaborative settings where the human and robot have the same objective, and the human is uncertain about the robot's type (i.e., the robot's internal state). We extend a recursive combination of Bayesian Nash equilibrium and the Bellman equation to solve for optimal robot policies. Interestingly, we discover that it is not always optimal for collaborative robots to be transparent; instead, human and robot teams can sometimes achieve higher rewards when the robot is \textit{opaque}. In contrast to transparent robots, opaque robots select actions that \textit{withhold} information from the human. Our analysis suggests that opaque behavior becomes optimal when either (a) human-robot interactions have a short time horizon or (b) users are slow to learn from the robot's actions. We extend this theoretical analysis to user studies across $43$ total participants in both online and in-person settings. We find that --- during short interactions --- users reach higher rewards when working with opaque partners, and subjectively rate opaque robots as about equal to transparent robots. See videos of our experiments here: \url{https://youtu.be/u8q1Z7WHUuI}}

\keywords{Human-Robot Cooperation, Expectation and Intention Understanding, Legibility, Game Theory}



\maketitle
\section{Introduction}\label{sec:intro}

\begin{figure}[t]
\vspace{0.5em}
 \begin{center}
		\includegraphics[width=1\columnwidth]{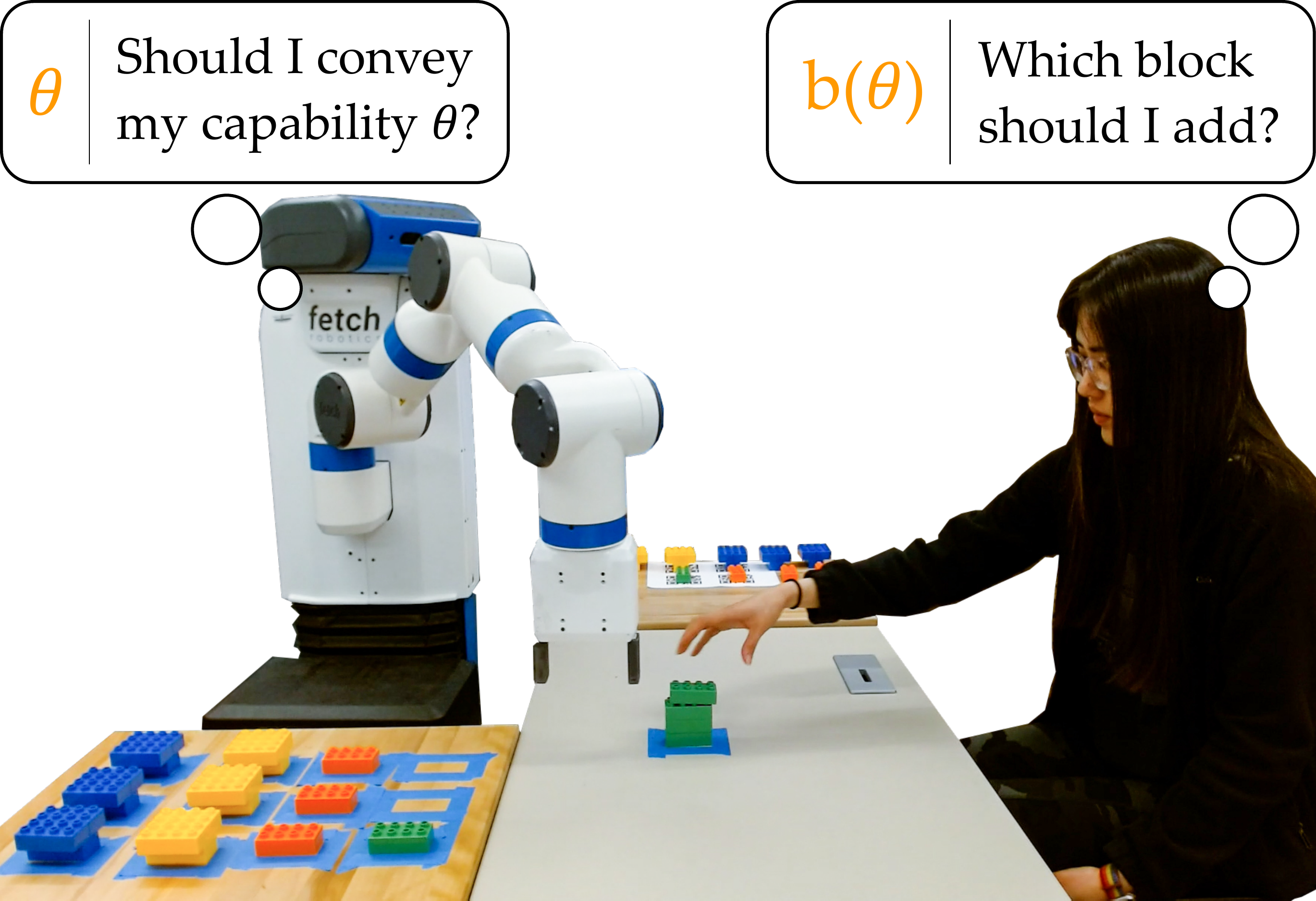}
		\caption{Collaborative block-stacking task where the human is uncertain about the robot's internal state $\theta$. Transparent robot actions help the human learn $\theta$ and decide what blocks to add to the tower. However, we find that the costs of this transparent behavior may outweigh its benefits}
		\label{fig:front}
	\end{center}
\end{figure}

Optimal robots select actions to maximize their objective function.
When robots collaborate alongside humans, maximizing this objective requires teamwork: the robot must reason about how the human interprets and reacts to the robot's behavior in order to seamlessly complete the overall task \cite{ijcai2021p61, foerster2018learning, Biyik2022, hakli2017cooperative}.
In this paper we focus on settings where the robot and the human share the same objective function (i.e., the robot and human are working together to perform a task), but the human does not know exactly how the robot will behave. 
This applies to factory floors, manufacturing, and assembly contexts where everyday human workers must collaborate with robot partners \cite{sanneman2021state, parekh2023learning, onnasch2021taxonomy, MUKHERJEE2022102231}.
For example, imagine a person teaming up with a robot arm to build a block tower (see \fig{front}). 
Both the human and robot share the same objective: they are trying to maximize the tower's height without it falling over. 
But the human is not sure about the robot's capabilities. 
If the robot is \textit{capable} it can reach for any block and add it to the tower; on the other hand, if the robot is \textit{confused} it will only be able to add the closer, smaller blocks. 
Whether the robot is capable or confused affects the human's optimal decisions: when working with a capable robot the human should add larger blocks, but for a confused robot the human needs to add smaller blocks to keep the tower from becoming unstable and falling over.
Given this uncertainty, it seems intuitive that the robot's optimal behavior is to pick up blocks that reveal whether it is capable or confused.

In line with this intuition, today's approaches to human-robot interaction often assume that robot behavior should be \textit{transparent} (e.g., legible, explainable, understandable) \cite{hellstrom2018understandable, sebo2020robots, hoffman2019evaluating, vitale2018bemore}. 
Transparent robots take actions that purposely reveal their internal state. 
For instance, when reaching for a block on a cluttered table, a transparent robot will exaggerate its trajectory so that nearby humans can predict which block the robot is going to grab \cite{dragan2013legibility, dragan2015effects, bodden2018flexible}. 
Transparent motions are beneficial because they convey information to the human, and the human can then leverage this information to better coordinate with the robot \cite{habibian2022encouraging, wu2021too, roncone2017transparent, kim2006who}. 
But transparent behavior also comes at a cost. 
Consider our example of reaching for a block: by exaggerating its trajectory the robot takes longer to get to the block and complete the task. 
Going one step further, the human teammate may require multiple interactions before they correctly interpret what the robot is trying to convey and update their own behavior in response.

In this paper we explore the situations where transparent behavior is optimal for human-robot teams. We build on related works to introduce \textit{opacity} as the opposite of transparency: opaque robots select actions that withhold information from the human. To determine whether it is optimal for robots to withhold information and select opaque behaviors, or to convey information and select transparent behaviors, our insight is that:
\begin{center}\vspace{0.4em}
\textit{We can formulate collaborative interactions where the human is uncertain about the robot internal state as a two-player stochastic Bayesian game.}
\vspace{0.4em}
\end{center}
We develop an algorithmic framework to solve these games and obtain optimal robot policies for each internal state (i.e., for each \textit{type} of robot). Interestingly, we find that --- under some conditions --- the optimal policy is the \textit{same} for every robot type and the robot's resulting behavior is opaque to the human. Return to our motivating example in \fig{front}. Although we might have expected the capable robot to stack large blocks and the confused robot to stack small blocks, we will prove that under some conditions the human-robot team actually has a higher expected reward if both robots always build the smaller tower. Put another way, when the human and robot act optimally the robot is opaque, and does not take actions to convey to its capabilities to the human.

Overall, we make the following contributions:

\p{Formalizing Opacity} We capture settings where the human and robot have the same payoff and the human is uncertain about the robot's type as stochastic Bayesian games. Within this context we build on prior works to define fully and rationally opaque behavior. 

\p{Proving when Opacity is Optimal} We extend a recursive combination of Bayesian Nash Equilibrium and the Bellman equation to find optimal robot policies. We then show that these optimal policies can be opaque. Our analysis and simulations suggests that it is more likely for opaque behavior to be optimal when (a) interactions have a short time horizon and (b) humans are slow to learn from robot actions.

\p{Measuring User Responses to Opaque Robots} We conduct online and in-person user studies with a total of $43$ participants. The major aim of these experiments is to test our theoretical analysis and see whether opaque robot behavior can result in higher human rewards during shorter interactions. We also conduct surveys to assess how users subjectively respond to collaborations with transparent or opaque partners.


\section{Related Work} \label{sec:related}

\noindent \textbf{Human-Robot Collaboration.} We consider cooperative settings where a human and robot are working together towards a common objective. This presents challenges from both human and robot perspectives: the robot must anticipate the human’s behavior, and the human needs to learn about the robot so that they can reliably interact \cite{clodic2021implement}. Developing mutual understanding between the human and robot is therefore essential for efficient collaboration \cite{clodic2017key, glasauer2010interacting, nikolaidis2015efficient}. Within our paper, we focus on scenarios where the human is uncertain about the robot (e.g., the human does not know what the robot has learned or how it will interact). Prior works have developed a variety of explicit communication models that robots can follow to convey information to humans \cite{habibian2023review}. For example, robots can take advantage of augmented reality headsets or visual interfaces to directly indicate their intent to a human partner \cite{walker2018communicating, christie2023limit}. But implicit communication and joint actions are also an effective tool for facilitating mutual understanding in human-robot interaction \cite{frijns2023communication}. For example, the way that a robot moves its arm could make it clear where it intends to go or what object it is planning to reach. In practice, implicitly communicating with robot actions presents a tradeoff: exaggerating how the robot moves its arm might make its goal clear, but it could also cause the robot to take longer to complete the task. We will propose a game-theoretic approach to resolve this problem and determine when the robot should take actions that convey information to its human partner, and when the robot should take actions that efficiently complete the shared task.

\noindent \textbf{Transparent Robots.} Prior work on human-robot interaction often assumes that robots should be \textit{transparent} (e.g., legible or explainable) \cite{hellstrom2018understandable, sebo2020robots, hoffman2019evaluating, dragan2013legibility}. Transparent robots actively and intentionally reveal their internal state to nearby humans so that these humans have a more accurate estimate of the robot's intent. Here we specifically focus on transparent robot actions (e.g., motions). Recent research demonstrates that robots can exaggerate their actions to communicate their goal \cite{dragan2015effects}, indicate their intended trajectory \cite{bodden2018flexible}, and express whether or not they are capable of performing a task \cite{kwon2018expressing}. When applied to collaborative human-robot teams where both agents are working together to complete a shared task, experiments suggest that transparent robot motion improves overall team performance
\cite{habibian2022encouraging, wu2021too, roncone2017transparent}.

Although transparency is often perceived as a benefit to human-robot interaction, we explore the opposite perspective: is it ever optimal for robots to \textit{hide} their internal state and intentionally \textit{withhold} information from the human?

\p{Deceptive Robots} Robot actions can be deceptive or misleading. For instance, by initially moving towards the wrong goal a robot can convince the human that this goal is what the robot actually wants --- even if the robot has another target in mind \cite{sreedharan2022obfuscatory, dragan2015deceptive}. We are not interested in explicitly encouraging deceptive actions; instead, we study situations where this behavior emerges \textit{naturally} as part of the robot's optimal policy. Towards this end we will formulate human-robot interaction as a two-player collaborative game \cite{nikolaidis2017game, bansal2022bayes, peters2020inference, albrecht2016belief}. By solving this game we seek to identify whether optimal robot behavior reveals or withholds information about the robot's latent state. When solving similar games, recent research has found that it can be optimal for robots to take actions that influence humans \cite{sagheb2022towards, sadigh2016planning, willi2022cola} or mislead users \cite{losey2019robots}.  

In these prior works the human and robot are \textit{competitors}: the robot has a different objective than the human. For instance, in \cite{sagheb2022towards} an autonomous car and human driver are competing to cross the intersection first, and the autonomous car takes misleading actions to influence the human driver to yield. By contrast, in our paper the human and robot are \textit{collaborative}: both agents share the exact same reward function and have no incentive to mislead one another.


\section{Problem Formulation} \label{sec:problem}

We consider industrial applications where a robot is working alongside a human partner. More specifically, we focus on \textit{collaborative} interactions where both the human and robot share the same reward function (i.e., both agents get the exact same payoff at every timestep). While we assume that the human and robot know this reward function (i.e., both agents know the task), the human is \textit{uncertain} about the robot's \textit{type}. Here type $\theta$ captures latent information observed only by the robot. For instance, consider our running example where a human is building a block tower with a robot arm: the robot can either reach any block (type capable) or the robot can only reach for nearby blocks (type confused). The human does not initially know the robot's type $\theta$ and the robot must decide whether to take actions that reveal its type during interaction.

\p{Interaction} Let $s \in \mathcal{S}$ be the system state, let $a_\mathcal{R} \in \mathcal{A}_\mathcal{R}$ be the robot action, and let $a_\mathcal{H} \in \mathcal{A}_\mathcal{H}$ be the human action. The system state transitions using deterministic dynamics: 
\begin{equation} \label{eq:P1}
    s^{t+1} = f(s^t, a_\mathcal{H}^t, a_\mathcal{R}^t)
\end{equation}
where both human and robot actions affect the system state. At each timestep the human and robot act simultaneously. Neither agent goes first: both select their actions $a_\mathcal{R}^t$ and $a_\mathcal{H}^t$ without knowing what action the other agent is taking. The interaction ends after a total of $T$ timesteps.

\p{Robot Type} Throughout the interaction the robot has a fixed type $\theta$. Within our experiments \textit{type} refers to the robot's level of capability (e.g., how quickly the robot can move or how effectively the robot can grasp objects). More generally, $\theta$ is latent information known only by the robot. Let there be $N$ possible types of robots. In our user studies we focus on cases where $N=2$ and there two possible types of robots (to better measure opacity and transparency), but the formulation we present extends to settings with an arbitrary number of robot types. At the start of each interaction $\theta$ is sampled from the prior $P(\theta)$. The human knows this prior distribution $P(\theta)$ but not the robot's current type $\theta$.

\p{Belief} The human updates their estimate of the robot's type during the interaction based on the robot's behavior. Let $b^{t+1}(\theta) = P(\theta \mid s^{0:t}, a_\mathcal{R}^{0:t})$ be the probability that the robot is type $\theta$ given that we have visited states $s^{0:t}$ and the robot has taken actions $a_\mathcal{R}^{0:t}$ up to the current timesetp $t$. Similar to prior work \cite{ziebart2008maximum}, we assume that the human updates this belief using Bayesian inference:
\begin{equation} \label{eq:P2}
    b^{t+1}(\theta) \propto b^t(\theta) \cdot P(a_\mathcal{R}^t \mid s^t, \theta)
\end{equation}
where $b^0(\theta) = P(\theta)$ is the known prior over the robot's type. The likelihood function $P(a_\mathcal{R} \mid s, \theta)$ expresses --- from the human's perspective --- how likely the robot is to take action $a_\mathcal{R}$ at state $s$ given that the robot is type $\theta$. Our approach is not tied to a specific choice of this likelihood function. In our simulations (Section~\ref{sec:sims}) we will test different models for $P(a_\mathcal{R} \mid s, \theta)$.

\p{Reward} At each timestep the collaborative human and robot receive the same reward $r(s)$. This reward function captures the task that the agents are trying to complete (e.g., the height of the tower the human and robot are building). Both agents know this reward function and know that the other agent shares the same reward. Summing reward at each timestep gives the total reward across the entire interaction: $\sum_{t=0}^T r(s^t)$.

\p{Stochastic Bayesian Game} The human and robot want to complete the task and maximize their reward across the interaction. If we model the human as a rational agent, this problem is an instance of a two-player stochastic Bayesian game where the human is uncertain about the robot's type \cite{albrecht2016belief}. We highlight two non-standard aspects of our game-theoretic formulation. First, the game is entirely collaborative: the agents always receive the same payoffs. Second, the human updates their belief according to \eq{P2}, and different humans may leverage different likelihood functions when updating this belief.

\p{Policies} Within a stochastic Bayesian game the human and robot each have policies. The human's policy $a_\mathcal{H} = \pi_\mathcal{H}(s, b)$ maps the current state and belief to actions, and the robot's policy $a_\mathcal{R} = \pi_\mathcal{R}(s, b, \theta)$ maps the state, belief, and type $\theta$ to robot actions. Remember that the robot knows its type $\theta$. In practice, robots of different types may take different actions given the same state and belief. Returning to our running example of building a tower: at a given $(s, b)$ a capable robot could reach for the larger block while the confused robot might move for the closer, smaller block.

In Section~\ref{sec:method} we will solve for the optimal robot policy. This policy $\pi_\mathcal{R}(s, b, \theta)$ will maximize the human's and robot's expected reward across an interaction.


\section{Should Collaborative Robots be Opaque?} \label{sec:method}

In this section we develop a solution to our stochastic Bayesian game to identify the robot's optimal policy.
Because the human and robot share the same reward function (and are therefore collaborating on a common task), it may seem reasonable to expect that the optimal robot behavior is \textit{transparent}.
Recall that transparent motions intentionally improve the human's estimate of the robot's latent state \cite{dragan2013legibility, hellstrom2018understandable}.
Returning to our motivating example in \fig{front}, we might expect the capable robot to take actions that reveal it can reach for larger blocks, and the confused robot to take actions that demonstrate it is limited to the nearby, smaller blocks. 
However, in this section we theoretically prove that transparency is not always optimal. 
We start by introducing formal definitions for \textit{opacity} when the robot is interacting with rational and irrational humans (Section~\ref{sec:method1}). Next, we derive a game-theoretic approach for finding \textit{optimal} robot behavior in collaborative games (Section~\ref{sec:method2}). Finally, we provide an example to prove that --- within our problem setting --- there exist cases where it is optimal for robots to be opaque and hide their capabilities (Section~\ref{sec:method3}).

\subsection{Formalizing Robot Opacity} \label{sec:method1}

We start by introducing two definitions for opacity within stochastic Bayesian games where the human is uncertain about the robot's type $\theta$. These definitions are consistent with prior work on robot legibility and transparency \cite{chakraborti2019explicability}.
Our first definition makes no assumptions about the human's policy, and applies when the robot is interacting with \textit{any} human partner. Our second definition assumes that the human acts \textit{rationally}: i.e., the human chooses actions to maximize the expected cumulative reward. For both definitions we reason about opacity in terms of the human's belief at the end of the interaction, $b^T$. Recall that $b(\theta)$ from \eq{P2} is the human's belief over the robot's type $\theta$, and the interaction has a total of $T$ timesteps. 
Intuitively, a transparent system causes the human to reach different beliefs when interacting with different types of robots, so that $b^T(\theta_1) \neq b^T(\theta_2)$ \cite{macnally2018action}. By contrast, an opaque system causes humans to converge to the same final belief regardless of the robot's actual type, so that $b^T(\theta_1) = b^T(\theta_2)$. More formally, we define Fully Opaque and Rationally Opaque robot's below:

\p{Fully Opaque} Let $(s^0, b^0)$ be the initial system state and prior over the robot type. Define $(s^0, b^0)$ as fully opaque if --- no matter which type $\theta \in \Theta$ the robot actually is --- the human's final belief $b^T(\theta)$ is equivalent. 

\smallskip

As a example, consider a robot where $a_\mathcal{R} = \pi_\mathcal{R}(s, b, \theta)$ is always zero; i.e., a robot that always takes action $a_\mathcal{R} = 0$. The human cannot distinguish what type of robot they are working with regardless of what policy $\pi_\mathcal{H}$ the human chooses. Accordingly, the human's final belief $b^T(\theta_i) = b^T(\theta_j)$ for any choice of $i$ and $j$. We define this robot as \textit{fully opaque} because, no matter what the human does, they always converge to the same final understanding of the robot's type.

\p{Rationally Opaque} Let $(s^0, b^0)$ be the initial system state and prior, and assume the human takes actions to maximize their expected total reward in the stochastic Bayesian game (i.e., the human acts optimally). Define $(s^0, b^0)$ as rationally opaque if --- no matter which type $\theta \in \Theta$ the robot actually is --- the rational human's final belief $b^T(\theta)$ is identical.

\smallskip

The difference between \textit{fully opaque} and \textit{rationally opaque} is our assumption about the human's policy. In both cases the robot does not reveal any information about its type over the course of interaction. But for rationally opaque the human is \textit{constrained} to always take optimal actions, while for fully opaque the human can take random actions to perturb the system. As we will show, a robot starting at $(s^0, b^0)$ may be rationally opaque \textit{but not} fully opaque (i.e., an irrational human could take random actions that cause the robot to reveal its underlying type).

\subsection{Identifying Optimal Behavior} \label{sec:method2}

Now that we have defined opacity, we will next determine if it is ever optimal for robots to be opaque. In Section~\ref{sec:problem} we formulated the human and robot as agents in a two-player stochastic Bayesian game. In this subsection we present an algorithm for finding optimal human and robot policies $\pi_\mathcal{H}$ and $\pi_\mathcal{R}$ under this game-theoretic setting. Next, in Section~\ref{sec:method3} we will apply this algorithm to example scenarios and demonstrate that the robot's optimal behavior in these settings can be fully opaque or rationally opaque.

\p{Augmented State} The human's policy $\pi_\mathcal{H}(s, b)$ and robot's policy $\pi_\mathcal{R}(s, b, \theta)$ depend on the system state $s$ and the human's belief $b$. The state transitions according to \eq{P1} and the belief transitions according to \eq{P2}. Without loss of generality, we combine these two equations into a single dynamics:
\begin{equation} \label{eq:M1}
    \big(s^{t+1}, b^{t+1}\big) = F\big(s^t, b^t,  a_\mathcal{H}^t, a_\mathcal{R}^t\big)
\end{equation}
where $(s, b)$ is the \textit{augmented state} and $F$ is the augmented dynamics under which the state and belief evolve. Note that $F$ is not a new equation: we are evaluating \eq{P1} and \eq{P2} to find the next state-belief pair $(s^{t+1},b^{t+1})$ given that the human takes action $a_\mathcal{H}$ and the robot takes action $a_\mathcal{R}$.

\p{Harsanyi-Bellman \textit{Ad Hoc} Coordination} Recent research finds optimal policies for stochastic Bayesian games through a recursive combination of Bayesian Nash equilibrium and the Bellman equation. This method is referred to as Harsanyi-Bellman \textit{Ad Hoc} Coordination (HBA) \cite{albrecht2015game}. Define $V(s,b)$ as the \textit{value} of the augmented state $(s,b)$, i.e., the total reward of starting in $(s,b)$ and acting optimally thereafter. Because both of the agents in our setting have common payoffs, and because there is only uncertainty about the robot's type, HBA applied to our context reduces to:
\begin{multline}\label{eq:M2}
    V\big(s, b\big) = r\big(s\big) + \\ \underbrace{\max_{a} \sum_{i = 1}^N b(\theta_i) \cdot V\Big(F(s, b, a_\mathcal{H}, a_{\mathcal{R}, i})\Big)}_{\text{Bayesian Nash Equilibrium at } (s,b)}
\end{multline}
where $a_{\mathcal{R}, i}$ is the action assigned to the $i$-th type of robot, and $a = (a_\mathcal{H}, a_{\mathcal{R},1}, a_{\mathcal{R},2}, \ldots, a_{\mathcal{R},N})$ includes the human's action and an action for each of the $N$ types of robots. We will denote the action $a$ that maximizes the underlined portion of \eq{M2} as $a^*=(a^*_\mathcal{H}, a^*_{\mathcal{R},1}, a^*_{\mathcal{R},2}, \ldots, a^*_{\mathcal{R},N})$.

Overall, \eq{M2} is an instance of the Bellman equation \cite{russell2022artificial}. The value $V(s,b)$ is equal to the immediate reward at $s$ plus the maximum expected value of the next augmented state. This expectation is taken over the human's belief in the robot's type: recall that $b(\theta_i)$ is the probability --- from the human's perspective --- that the robot is type $i$. Within the underlined portion of \eq{M2} we find the Bayesian Nash Equilibrium at the current augmented state $(s,b)$. Specifically, we find an action $a^* = (a^*_\mathcal{H}, a^*_{\mathcal{R},1}, a^*_{\mathcal{R},2}, \ldots, a^*_{\mathcal{R},N})$ for the human and each type of robot such that $a^*$ maximizes the next value given the human's current belief $b$. Intuitively, the human is not sure which robot type they are dealing with: each type may take a different action $a_\mathcal{R}$, and the rational human identifies an action $a_\mathcal{H}$ that will maximize the expected value across robot types. In practice we can solve \eq{M2} by either (a) discretizing the states and actions and applying classical value iteration approaches \cite{russell2022artificial}, or by (b) leveraging recent algorithms that approximate the value function in continuous spaces \cite{lutter2021value}. In either case our output is the value $V$ at each augmented state $(s,b)$.

Now that we have $V(s,b)$ from \eq{M2} we will use this value to find optimal human and robot policies $\pi_\mathcal{H}$ and $\pi_\mathcal{R}$. Let $a^*$ be the Bayesian Nash Equilibrium at $(s,b)$:
\begin{equation}\label{eq:M3}
    a^* = \text{arg}\max_{a}\sum_{i = 1}^N b(\theta_i) \cdot V\Big(F(s, b, a_\mathcal{H}, a_{\mathcal{R}, i})\Big)
\end{equation}
where $a^* = (a^*_\mathcal{H}, a^*_{\mathcal{R},1}, \ldots, a^*_{\mathcal{R},N})$ assigns an action to each type of robot, so that robot type $\theta_i$ takes action $a^*_{\mathcal{R}, i}$. The optimal human and robot take their respective actions within this Bayesian Nash Equilibrium, such that:
\begin{equation} \label{eq:M4}
    \pi_\mathcal{H}(s,b) = a^*_\mathcal{H}, \quad  \pi_\mathcal{R}(s,b,\theta_i) = a^*_{\mathcal{R}, i} 
\end{equation}
Overall, Equations (\ref{eq:M2})--(\ref{eq:M4}) solve the stochastic two-player Bayesian game to find the optimal pair of policies for the human and robot. In practice, we recognize that actual human users may deviate from their optimal policy $\pi_\mathcal{H}$. Humans that exactly follow $\pi_\mathcal{H}$ are \textit{rational humans}: in our definition of \textit{rationally opaque} we assume that the human executes $\pi_\mathcal{H}$. We also emphasize that, when solving Equations (\ref{eq:M2})--(\ref{eq:M4}) for the robot's optimal policy $\pi_\mathcal{R}$, we have assumed the human partner acts rationally.

\subsection{Proving Opaque Behavior can be Optimal} \label{sec:method3}

By leveraging our modified HBA algorithm from Section~\ref{sec:method2} we can find the robot's optimal policy and determine whether it is ever optimal for robots to be opaque.
Here we apply Equations (\ref{eq:M2})--(\ref{eq:M4}) to a simulated $1$-DoF human-robot team\footnote{See the complete implementation of this example here: \url{https://github.com/VT-Collab/opaque}}. 
We first empirically demonstrate that it \textit{can} be optimal for collaborative robots to be fully opaque. Next, we explore the distinction between fully and rationally opaque, and prove that robots which are rationally opaque \textit{may not be} fully opaque.

\begin{figure}[t]
    \vspace{0.5em}
	\begin{center} \includegraphics[width=1\columnwidth]{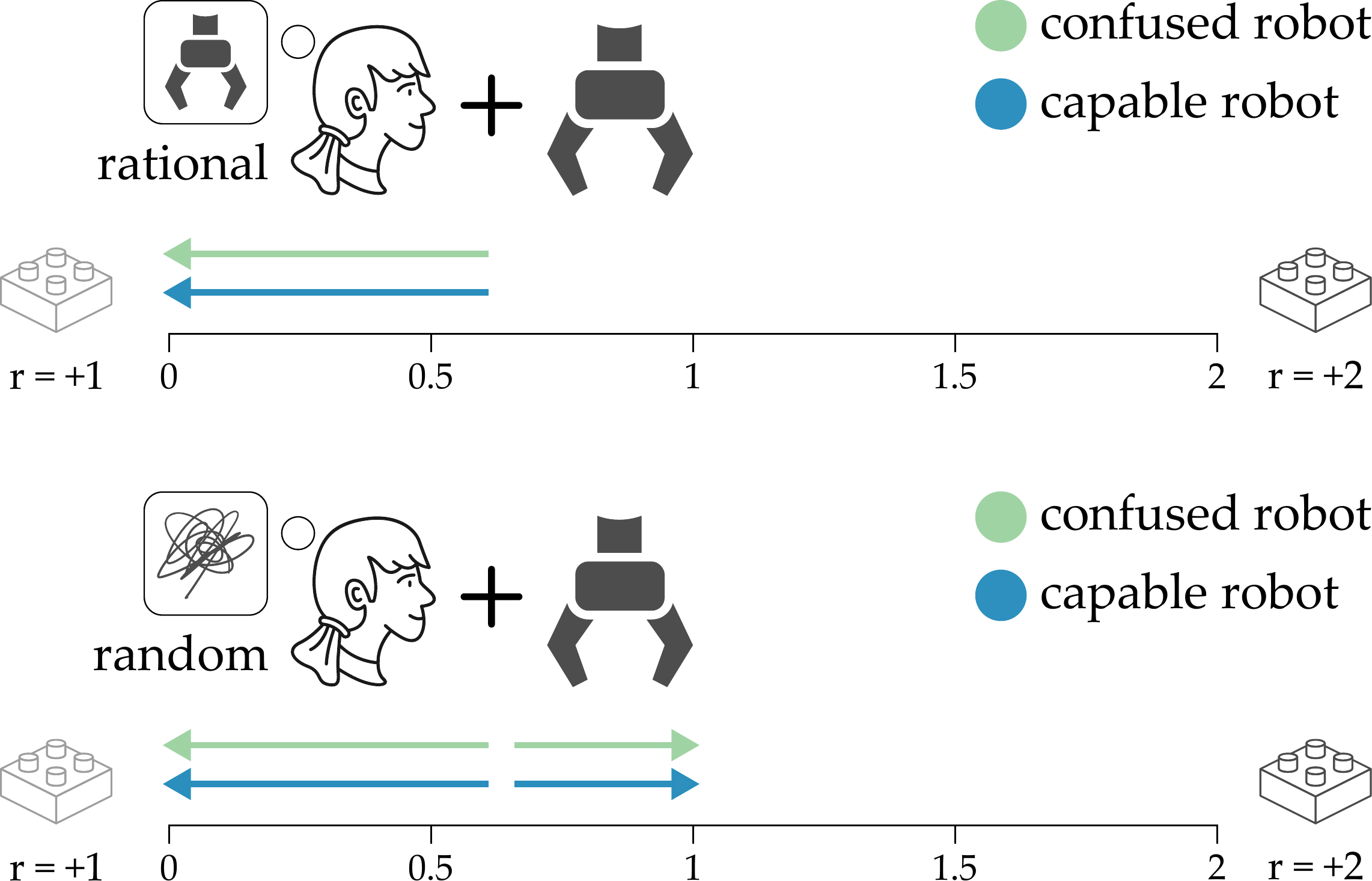}
    \put(-220,130){(a)}
    \put(-220,55){(b)}
		\caption{Example of an optimal, fully opaque robot. The system starts at position $s = 0.6$ and prior $b^0 = 0.2$. The robot has two types $\theta$: confused and capable. The confused robot can only move towards the left block. (a) Optimal human and robot solve this stochastic Bayesian game. (b) Optimal robot is paired with a human that takes random actions. Regardless of the human's actions, both capable and confused robots always move towards the left block. Hence, the robot is \textit{fully opaque}, and the human cannot infer $\theta$ from the optimal robot's actions}
		\label{fig:example1}
	\end{center}
\end{figure}

\p{Example Problem} Consider a $1$-DoF version of our motivating example where the human and robot are trying to reach for a block (see \fig{example1}). The state $s$ is the position of the human-robot team. This state is bounded between $[0,2]$, where ${s=0}$ is the position of the block closer to the robot and ${s=2}$ is the position of the farther block. At each timestep the human can take actions $\mathcal{A}_\mathcal{H} = \{-0.2, 0, +0.2\}$ to reach left or right. There are two types of robots: a \textit{capable} robot ($\theta_1$) which can reach for either block, and a \textit{confused} robot ($\theta_2$) that can only move for the block at ${s=0}$. The capable robot's action set is $\mathcal{A}_{\mathcal{R},1} = \{-0.1, +0.1\}$ and the confused robot's action set is $\mathcal{A}_{\mathcal{R},2} = \{-0.1\}$. The human increases their belief $b$ that the robot is capable if they observe $a_\mathcal{R}=+0.1$. The game ends after a total of $T=5$ timesteps. The human and robot receive reward $r = +1$ if they end the game at state $s=0$ (the closer block) $r=+2$ if they end the game at $s=2$ (the farther block). At all other states the reward is $r=0$.

 \begin{figure*}[t]
	\begin{center}		
    \includegraphics[width=2\columnwidth]{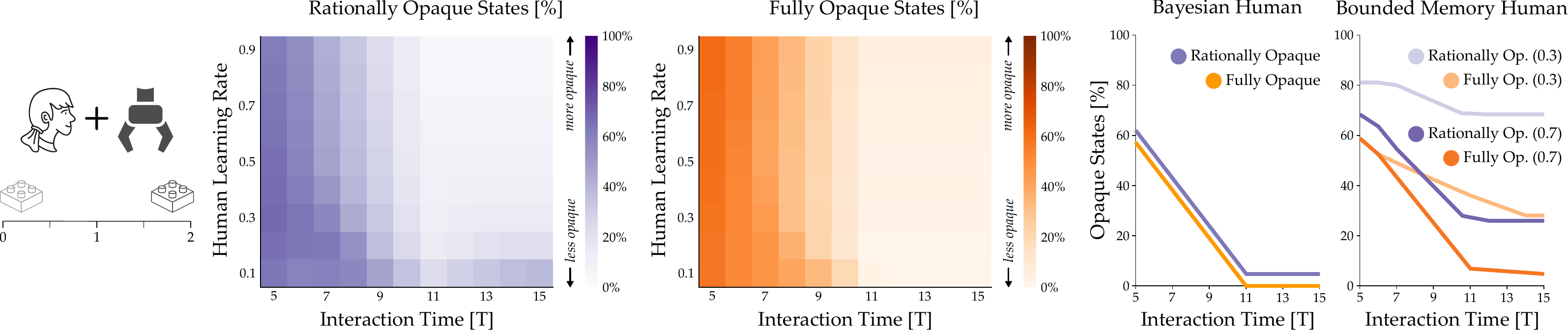}
    \put(-435,86){(a)}
    \put(-370,86){(b)}
    \put(-138,86){(c)}
    
		\caption{Simulation results from our $1$-DoF Environment. (a) The human and robot collaborated to reach a goal; the confused robot could only go left while the capable robot could help reach right or left. For each plot we sampled all start states and priors and then calculated the percentage of those augmented states which were opaque; e.g., $50\%$ opaque means that for half of the initial augmented states it was \textit{optimal} for the robot to withhold its type $\theta$ from the human. (b) We varied the human's learning rate and the total number of timesteps in each interaction. A higher learning rate indicated that the human uncovered $\theta$ more quickly when the actions for each robot type diverged. (c) We also tested a human that used Bayesian inference to update their belief and two bounded memory humans (with learning rates of $0.3$ and $0.7$) that forgot what they had learned after each timestep}
		\label{fig:sim1d}
	\end{center}
    \vspace{-0.5em}
\end{figure*}

\begin{figure*}[t]
	\begin{center}		
    \includegraphics[width=2\columnwidth]{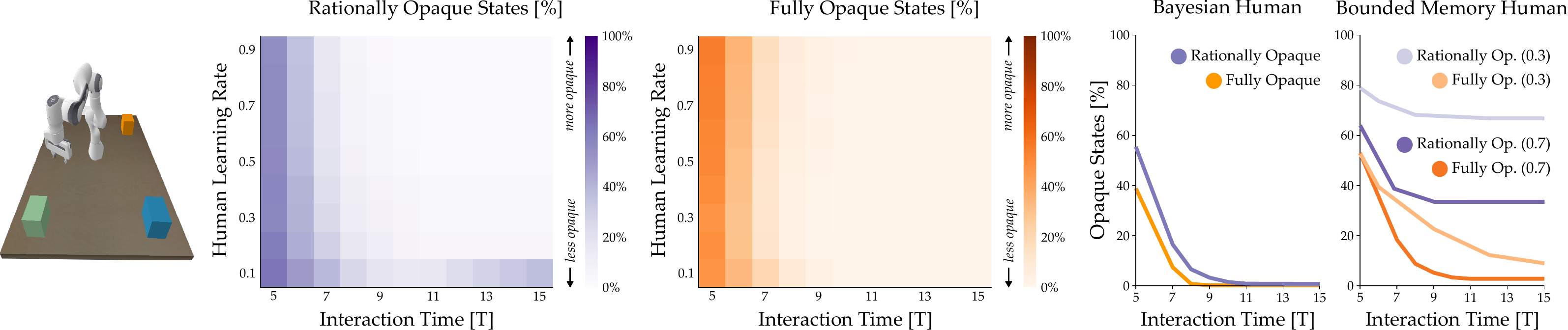}
    \put(-435,86){(a)}
    \put(-370,86){(b)}
    \put(-138,86){(c)}
		\caption{Simulation results from our robot arm environment. (a) Humans shared control with a robot arm to reach for goals on the table; the capable robot could go towards any goal while the confused robot could only move down and left. The format of our results follows \fig{sim1d}. (b) The number of opaque states decreases as the interaction time increases. (c) The number of opaque states also decreases as the human learns more quickly. Note that the Bayesian human is an \textit{ideal} user that can infer the robot's type from a single timestep; i.e., this human model learns $\theta$ as efficiently as possible. When compared to this ideal human, it is more likely for opaque behavior to be optimal when the robot is collaborating with a forgetful user that follows the bounded memory model. Overall, our results show that opaque behavior is more likely to be optimal during short interactions with suboptimal humans}
		\label{fig:sim2d}
	\end{center}
\end{figure*}

\p{Known Type} Imagine that the robot's type $\theta$ is public knowledge. If the human interacts with a capable robot, both the human and robot should move right at every timestep ($a_\mathcal{H}= {+0.2}$ and $a_\mathcal{R}= +0.1$) to reach the farther block (reward $r=+2$). Alternatively, if the human knows they are interacting with the confused robot and they start at a state $s < 1.5$, then they cannot reach the farther block and should move left for the closer block ($a_\mathcal{H}= {-0.2}$ and $a_\mathcal{R}= -0.1$). The human's optimal actions depend on the robot's type $\theta$. It may therefore seem optimal for the robot to reveal its type so that the human can determine which block to aim for. As we will show, however, this is not always the case.

\p{Optimal Robots can be Fully Opaque} Let initial system state be $s^0 = 0.6$ and let the prior be $b^0(\theta_1) = 0.2$ (i.e., the human is $20\%$ sure the robot is capable). We solve Equations (\ref{eq:M2})--(\ref{eq:M4}) to find the optimal robot policy $\pi_\mathcal{R}$. We then pair this optimal robot with two different humans: a \textit{rational} human that follows the optimal human policy $\pi_\mathcal{H}$, and a \textit{random} human that can take any action (see \fig{example1}). We find that --- no matter what action the human takes --- the optimal action for both types of robots is to move left and reach for the closer goal. Put another way, when starting at this $(s^0, b^0)$ both robot types always take action $-0.1$. Rational or random humans cannot distinguish the robot's type: at the end of the interaction the human's belief is $b^T(\theta_1)=0$ after working with both capable and confused robots. Accordingly, given the initial augmented state $s^0 = 0.6$ and $b^0(\theta_1) = 0.2$ the robot's \textit{optimal} behavior is \textit{fully opaque}. Moving for the closer block results in higher team reward, even though it does not convey information to the human.


\p{Optimal Robots can be Rationally Opaque} We next prove that an optimal robot may be rationally opaque \textit{but not} fully opaque. Let the initial state be $s^0 = 1.0$ with prior $b^0(\theta_1) = 0.2$ (i.e., the $1$-DoF system starts farther to the right than before). When the optimal robot interacts with a rational human the system again reaches for the closer block; both capable and confused robots always take action $a_\mathcal{R} = -0.1$, and the rational human's final belief is identical across both types of robots. But this changes when the human is free to take any action. If the random human takes action $a_\mathcal{H}=0.2$ the capable robot switches direction to go towards the farther goal and obtain reward $r=+2$. The random human's final belief is $b(\theta_1) = 0.4$ when interacting with the capable robot and $b(\theta_1) = 0$ with the confused robot. Hence, for this initial state the robot's optimal behavior is only \textit{rationally opaque}.


\section{What Conditions Lead to Opaque Robots?}
\label{sec:sims}

In Section~\ref{sec:method} we developed a method for solving our stochastic Bayesian game and identifying optimal robot behavior. We also showed that opaque robot behavior can be optimal in certain problem settings.
In this section we analyze \textit{what types of problem settings} lead to optimal, opaque robots.
We conduct controlled experiments with simulated humans. 
We hypothesize that the duration of the interaction and the speed of the human's learning will determine whether it is optimal for robots to be transparent or opaque.
Accordingly, we vary the time horizon of the interaction, the human's learning rate, and the way the human learns from the robot's behavior. 
Across two simulated environments, we observe that shorter interaction times and lower learning rates result in a higher number of optimally opaque states.

\p{Environments} Our simulated environments are shown in \fig{sim1d} and \fig{sim2d}. All code for replicating these environments and reproducing our simulations is available online\footnote{See \url{https://github.com/VT-Collab/opaque}}. The $1$-DoF human-robot team matches the example from Section~\ref{sec:method3} with one slight difference: now the human actions $\mathcal{A}_\mathcal{H} = \{-0.1, 0, +0.1\}$ have the same magnitude as the robot actions. As a reminder, in this $1$-DoF setting the human and robot are collaborating to reach one of the goals. One type of robot can reach either goal, and the other type of robot can only move left.

We extend this environment to create a \textbf{robot arm} simulation (see \fig{sim2d}). In accordance with our motivating example of assembling a tower, the human and robot must collaborate to reach blocks (e.g., goals) around the table. There are three different goals on the table; the confused robot ($\theta_1$) can only move down or to the left, while the capable robot ($\theta_2$) can autonomously move in any direction. Goals that are farther away from the robot's base have a higher reward. However, the confused robot may not be able to coordinate with the human to reach these goals, and thus the human needs to determine the robot's type to figure out which goal to aim for.

\p{Procedure} For each environment, time horizon, and simulated human we first solve Equations (\ref{eq:M2})--(\ref{eq:M4}) to find the optimal robot policies. We then sample all the discrete states and priors, and test whether the augmented start states $(s^0, b^0)$ are fully opaque, rationally opaque, or transparent. In what follows we report (a) the percentage of \textit{rationally opaque} start states and (b) the percentage of \textit{fully opaque} start states. Our overall results are summarized in \fig{sim1d} and \fig{sim2d}.

\subsection{Varying Interaction Time} \label{sec:S1}

The human-robot interaction ends after $T$ total timesteps. 
To understand how the duration of the interaction affected the optimality of opaque behaviors, we held the simulated human constant and varied the interaction time $T$.
When interactions are short (i.e., as $T \rightarrow 0$) we find that the robot's optimal behavior is \textit{opaque} for an increasing number of initial conditions. 
Consider the $x$-axes of \fig{sim1d} and \fig{sim2d}: the percentage of rationally opaque and fully opaque states \textit{increases} as $T$ \textit{decreases}.
Conversely, when interactions are long (i.e., as $T \rightarrow \infty$) the robot's optimal behavior becomes \textit{transparent}.
Again looking at the $x$-axes of \fig{sim1d} and \fig{sim2d}, the percentage of rationally opaque and fully opaque states \textit{decreases} as $T$ \textit{increases}.

To explain this result we highlight a trade-off between completing the task and communicating hidden information. Using the robot's actions to convey the robot's type $\theta$ requires additional timesteps: the robot must exaggerate its motion (and potentially move away from goals) to indicate $\theta$ to the human teammate. But when the interactions are short, this time is not available --- the robot must leverage all of its actions to directly complete the task and maximize the team's reward. Hence, we suggest that transparent behavior is more suitable for settings where the human and robot will work together for long periods of time, while opaque behavior is more likely to be optimal when the human and robot are collaborating across a single-shot task.

\subsection{Varying Human Learning Rate} \label{sec:S2}

The human updates their estimate $b(\theta)$ of the robot's type $\theta$ as they observe the robot's behaviors. 
For example, if the robot reaches for a block (i.e., a goal) that is farther away, the human should become more confident that the robot is capable of reaching these distant goals.
During our controlled simulations we adjusted how slowly or rapidly the simulated human's belief changed during a single timestep --- i.e., we adjusted the human's \textit{learning rate}.
For a learning rate of $0.1$ the next belief $b^{t+1}(\theta) = b^t(\theta) \pm 0.1$, and for a learning rate of $0.9$ the belief similarly updates in increments of $0.9$. 
Increasing the learning rate corresponds to a human that is more sensitive to differences in robot behavior; by contrast, lowering the learning rate towards zero corresponds to a human that does not update their estimate regardless of the robot's actions.
Looking at Figures~\ref{fig:sim1d} and \ref{fig:sim2d}, we find that the percentage of rationally opaque and fully opaque states \textit{decreases} as the learning rate \textit{increases}. 
When the simulated human learns more rapidly, \textit{transparent} behavior becomes optimal.
On the other hand, when the simulated human learns more gradually, it is optimal for robots to remain \textit{opaque}.

We explain this result in connection with the interaction length $T$ from Section~\ref{sec:S1}. When the learning rate increases the robot can more rapidly convey information to the human, and thus the robot does not need to devote as many actions and timesteps to communicate its type $\theta$.
At the other end of the spectrum, when the human learns gradually the robot must commit multiple actions to convey $\theta$, and during these actions the human's response is not necessarily aligned with the robot's capabilities.
At the extreme the human does not learn at all from the robot's actions: in this case, there is no advantage to transparency.
Our simulation results suggest that transparent robots are more likely to be optimal when the human is sensitive to the robot's behavior, and opaque behavior is more likely to be optimal when the human learns slowly or ignores the robot's actions.

\subsection{Varying How the Human Learns} \label{sec:S3}

We finally test two alternative human models. First, we simulate an \textit{ideal} human that leverages Bayesian inference to update their belief \cite{baker2011bayesian}. Here the human knows the robot's policy and treats $\pi_\mathcal{R}$ as the likelihood function in \eq{P2}. This ideal human learns as quickly as possible: at augmented states where $\pi_\mathcal{R}(s,b, \theta_i)$ and $\pi_\mathcal{R}(s,b, \theta_j)$ output different actions, the Bayesian human immediately distinguishes types $\theta_i$ and $\theta_j$. Second, we apply the bounded memory model \cite{nikolaidis2016formalizing, aumann1989cooperation} to simulate a human that only remembers the robot's most recent behavior (i.e., the robot's last action). This human updates their belief $b$ during each timestep at a fixed learning rate, and then resets $b$ to the prior between timesteps (i.e., this simulated human \textit{forgets} what they have learned). Our results for ideal and forgetful humans are shown on the right side of Figures~\ref{fig:sim1d} and \ref{fig:sim2d}. When the interaction lasts only a few timesteps $T$, even for an ideal human learner it is still optimal for the robot to select opaque behaviors. As the duration of the interaction \textit{increases} we again find that the percentage of opaque states \textit{decreases} for both the ideal and forgetful humans. We also find that bounded memory humans with a lower learning rate lead to an increased number of opaque states.

These results demonstrate that opaque robot behavior is not tied to a specific model of human learning: we find that opaque behavior can be optimal when the human learns as quickly as possible (\textit{Bayesian Human}) or when the human forgets previous actions (\textit{Bounded Memory Human}). In line with the results from Section~\ref{sec:S1} and Section~\ref{sec:S2}, we again observe the connection between interaction time $T$ and the speed of the human's learning. \textit{Transparent} behavior is more likely to be optimal as $T$ increases and the human learns more efficiently, while \textit{opaque} behavior is suitable during short interactions where the human is less sensitive to the robot's behavior.
In Section~\ref{sec:user} we will move beyond these simulated, controlled environments and test whether these optimization procedures and design principles apply with real-world users.


\section{User Studies} \label{sec:user}

 \begin{figure*}[t]
 \vspace{0.5em}
	\begin{center}		
    \includegraphics[width=2\columnwidth]{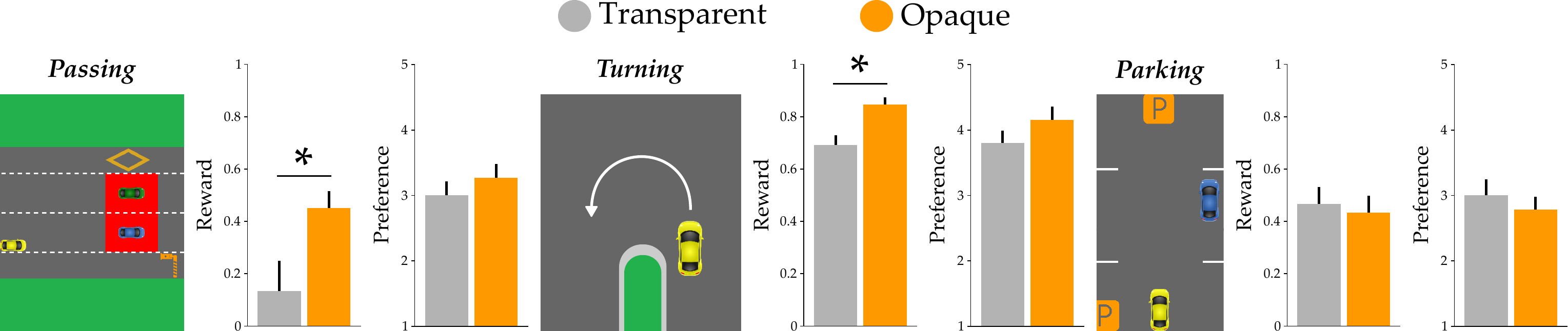}
		\caption{Task results from our online user study. Participants collaborated with a virtual agent to drive a car in \textit{Passing}, \textit{Turning}, and \textit{Parking} environments. Error bars show standard error and an $*$ denotes statistical significance ($p<.05$)}
		\label{fig:online}
	\end{center}
\end{figure*}


In this section we test how humans collaborate with opaque or transparent robot partners.
Our analysis from Section~\ref{sec:method} and simulations from Section~\ref{sec:sims} suggest that there are situations where opaque robot behaviors lead to better team performance.
In the following experiments we use Equations (\ref{eq:M2})--(\ref{eq:M4}) to solve the stochastic two-player Bayesian game and find optimal robot policies.
We consider single-shot driving environments and the tower-building task from \fig{front}.
Here the short interaction time $T$ results in optimal opaque behaviors --- i.e., when solving the system of equations the robot finds that opaque behaviors will lead to higher rewards.
But it is not clear whether \textit{actual humans} will follow this game-theoretic model, and if the behaviors that the robot predicts will result in higher rewards actually improve the team's performance.
In addition to the shared reward, we also monitor the human's subjective response.
Even if the human does receive higher rewards with an opaque robot, they may prefer working with transparent partners that take actions to convey their latent state.
To test both the team's performance and the human's perception we performed two separate user studies: an online user study with autonomous driving (Section~\ref{sec:user1}), and an in-person experiment with collaborative block stacking (Section~\ref{sec:user1}).
Below we first introduce our independent and dependent variables shared across both user studies, and then describe the protocol, results, and discussion for each specific experiment. Our code and user study results can be found here: \url{https://github.com/VT-Collab/opaque/}

\p{Independent Variables} During each trial one human interacted with one robot. There were two types of robots ($N=2$) that the human could interact with: these robots were not visually different, and the human had to infer the robot's type based on its actions. 
For ease of reference we will refer to these types as the \textit{confused} and \textit{capable} robots.
During each interaction the robot's actual type was randomized: in half of the trials participants worked with the \textit{confused} robot, and the other interactions were with the \textit{capable} robot. 
As such, participants were unsure about the robot's current type $\theta$ at the start of each interaction.

We varied the robot's algorithm to compare \textbf{Opaque} and \textbf{Transparent} robot policies. 
To find the \textbf{Opaque} policy we applied Equations (\ref{eq:M2})--(\ref{eq:M4}) from Section~\ref{sec:method2} and solved for the robot's optimal behavior $\pi_\mathcal{R}$. 
We then confirmed that this optimal policy $\pi_\mathcal{R}$ was fully opaque at the environments' start states according to the definition from Section~\ref{sec:method1}.
To find \textbf{Transparent} robot behavior we leveraged a baseline from related works \cite{dragan2013legibility, dragan2015effects}.
The \textbf{Transparent} algorithm modified the robot's reward to incentivize revealing actions; more formally, in \cite{dragan2013legibility, dragan2015effects} the robot assigns a bonus reward to actions that convey $\theta$ to the user.
We incorporated this bonus reward into our stochastic Bayesian game, and then solved for the optimal (and transparent) robot behaviors.
We confirmed that the \textbf{Transparent} robot took actions to communicate its intent and was not either fully or rationally opaque.

\p{Dependent Measures} During each interaction we measured the human-robot team's final \textit{Reward}. Here reward corresponded to the score in each environment: e.g., the distance the shared car traveled in the online study, or the height of the block tower in the in-person study. We normalized these rewards between $0$ and $1$ for consistent reporting. We also displayed the human's reward in real-time so that participants could track their own performance. 

Next, we measured the human's subjective \textit{Preference} for each robot algorithm. Preference was measured on a $1$-$7$ Likert scale, where higher scores indicated the human had a stronger preference for working with that robot algorithm.

\p{Hypotheses} Throughout our user studies we had two main hypotheses:
\begin{quote}
\p{H1} \textit{Humans will obtain more reward with \textbf{Opaque} partners than with \textbf{Transparent} partners.}
\end{quote}
\begin{quote}
\p{H2} \textit{Humans will prefer \textbf{Opaque} robots, even though these robots withhold information.}
\end{quote}

\subsection{Online: Sharing Control of an Autonomous Car} \label{sec:user1}

We first conducted an online survey where participants collaborated with a virtual robot to share control over a car. At each timestep the human clicked their input $a_\mathcal{H}$, the robot selected its action $a_\mathcal{R}$, and then the autonomous car moved using the combined action $a = a_\mathcal{H} + a_\mathcal{R}$.

\p{Experimental Setup} Participants teamed up with the robot to drive in three settings: passing, turning, and parking (see \fig{online}). In \textit{Passing} the team {received rewards for making lane progress, staying on the road, and avoiding a collision. In \textit{Turning} the reward was the car's velocity plus the total angle the car turned. Finally, in \textit{Parking} the team obtained a reward for either (a) parallel parking directly above the start position or (b) driving straight ahead to an open parking place. Each reward was normalized into a $0$-$1$ range.

Participants completed every scenario with \textbf{Opaque} and \textbf{Transparent} robots. The robot's type $\theta$ was randomized and counterbalanced --- in half of the interactions the robot was type confused, and in the other half the robot was type capable. Accordingly, there were four equally balanced experimental conditions: \textbf{Opaque} policies with confused and capable robots, and \textbf{Transparent} policies with confused and capable robots. We intentionally did not tell participants what type of robot they were currently interacting with. Under the \textbf{Transparent} policy, the robot took actions to try communicating its type to the human. By contrast, our analysis suggests that an optimal algorithm in this context should be \textbf{Opaque}: this method took the same assistive actions with both vehicles, so users did not have a way to determine if their robot partner was capable or confused.

Interactions lasted three timesteps ($T=3$). During each timestep we first showed the participant an image of the current state $s^t$ and prompted the user to select their action from a multiple choice menu (e.g., turning left, accelerating forward). After the user selected their action $a_\mathcal{H}^t$, the robot acted simultaneously with $a_\mathcal{R}^t$ and we showed an image of the next state $s^{t+1}$. We displayed the user’s reward throughout the game, and then at the end asked if the user ``preferred sharing control with this car.'' Overall, we measured the team’s objective performance (i.e., the reward), as well as the human’s subjective perception of their robot partner (i.e., the human’s preferences).

\p{Participants} We recruited $44$ anonymous participants. These participants were recruited using university mailing lists at Virginia Tech, the University of California, Berkeley, and the University of Illinois Urbana-Champaign. All users who agreed to participate in our study were first provided with the instructions. We then asked the participants two comprehension questions to check and see whether they had carefully read those instructions. Participants could go back and reread the instructions if needed to better answer the comprehension questions. In total, 30 of our 44 recruited users correctly answered both questions. These 30 participants continued on to the study, and we report their results below. For the remaining 14 users, the survey automatically terminated, and we did not collect their responses. Each participant performed $12$ driving interactions ($3$ scenarios, twice with \textbf{Opaque} and twice with \textbf{Transparent}). The order of presentation was randomized and counterbalanced so that half of the users started each scenario with the \textbf{Opaque} robot and the other half started with \textbf{Transparent}. Participants received a $\$5$ USD gift card after completing the survey.

\begin{figure}[t]
	\begin{center}		
    \includegraphics[width=1\columnwidth]{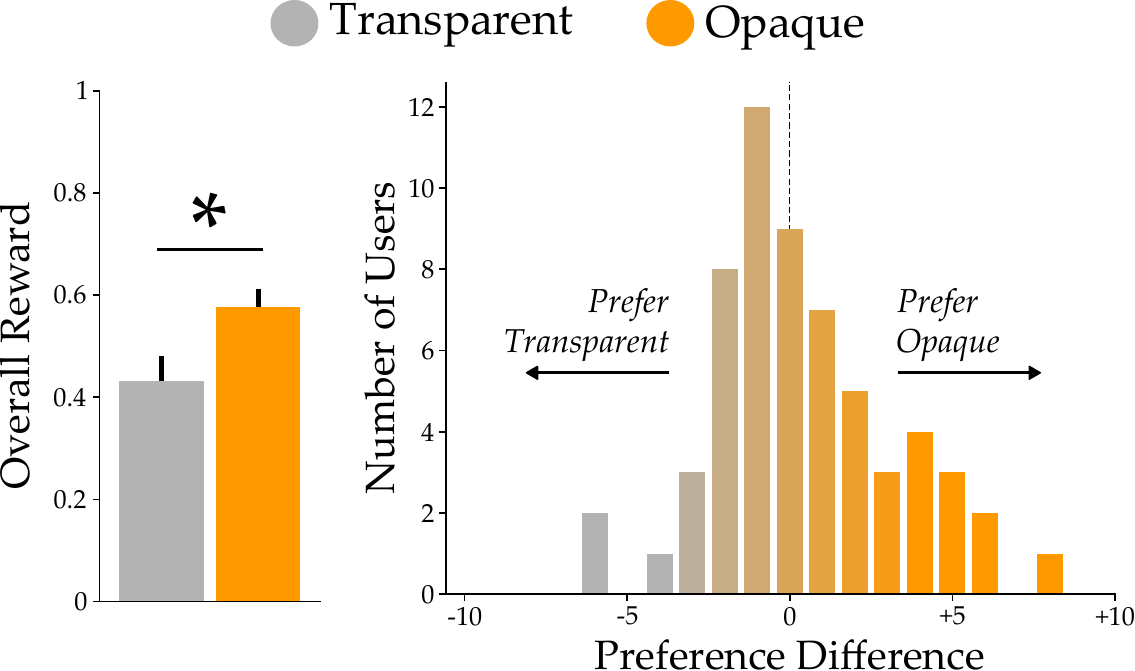}
    \put(-220,100){(a)}
    \put(-155,100){(b)}
		\caption{Overall results from our online user study. (a) The average reward across all three tasks. (b) $30$ participants completed each task twice with \textbf{Opaque} and twice with \textbf{Transparent}, resulting in $60$ pairs of datapoints. For each pair we subtracted the total preference scores with \textbf{Transparent} from the total preference scores with \textbf{Opaque}. Positive numbers indicate that individual user ranked \textbf{Opaque} as better than \textbf{Transparent}, and negative values indicate the opposite. We found that most users perceived the two algorithms as roughly equal (the Preference Difference was near zero). Error bars show standard error and an $*$ denotes statistical significance ($p<.05$)}
		\label{fig:online2}
	\end{center}
\end{figure}

\p{Results} Our results are summarized in  Figures~\ref{fig:online} and \ref{fig:online2}. This includes the objective outcomes of our experiment (the team’s reward) as well as the subjective outcomes (the human’s preference for each different robot policy). We first conducted a repeated measures ANOVA with three factors to identify significant main effects across types, policies, and tasks. We found that the team's overall reward was significantly higher when humans collaborated with \textbf{Opaque} partners as compared to when humans collaborated with \textbf{Transparent} partners ($F(1, 29) = 4.444$, $p = .044$). We also found that the human-robot team scored differently on different driving tasks ($F(2, 58) = 33.436$, $p < .001$). Breaking down these results using posthoc comparisons with Bonferroni corrections, we observe that in both Passing ($p<.001$) and Turning ($p<.001$) users reached higher rewards with \textbf{Opaque} partners. These results support hypothesis \textbf{H1} and suggest that real participants can collaborate more efficiently with optimal but opaque robots over short time scales.

On the other hand, the users’ perceptions of the robots were divided. Using repeated measures ANOVA with three factors to identify significant main effects across types, policies, and tasks. We found that the differences in preference scores were not statistically significant ($F(1, 29) = 0.872$, $p = .358$). We initially hypothesized that there might be a bi-modal split in how participants subjectively perceived the \textbf{Opaque} and \textbf{Transparent} algorithms. For example, if half of the users strongly preferred \textbf{Opaque} robots and the other half strongly preferred \textbf{Transparent} robots, then these two halves would cancel out when computing the average scores. But calculating each individual's preferences in \fig{online2}, we find that most users are roughly on the fence, and the overall distribution is unimodal. Hence, for hypothesis \textbf{H2} we find that online users did not clearly prefer either \textbf{Opaque} or \textbf{Transparent} robots. Overall, our statistical analysis here suggests that the robot’s algorithm did not have a statistically significant effect on the user’s preference, and we did not find a noticeable split between users. However, we recognize that this null effect is not conclusive, and we cannot claim that humans prefer \textbf{Transparent} and \textbf{Opaque} equally. It is possible that --- if we increased the number of participants --- we might find that the larger pool of users prefers one of these robots over the other.

\p{Discussion} When sharing control of the autonomous car, users needed to quickly coordinate with their robot partner to successfully complete the task (i.e., to pass, turn, or park). If the human and robot attempted to complete the task in different ways, their conflicting inputs resulted in lower rewards. Hence, our findings here are aligned with our simulation results from Section~\ref{sec:S1}, and suggest that transparent actions are not desirable when the human and robot must quickly collaborate. 

One limitation of this online study is that the participants may not have been invested in the outcomes. There were no consequences if the human-robot failed the task outside of a lower user score. For example, if humans were driving actual vehicles, we anticipate that their behavior may have been more conservative. 
To help address this limitation we next performed an in-person user study where participants worked directly with a physical robot arm.

\subsection{In-Person: Stacking Blocks with a Robot Arm} \label{sec:user2}

In our second user study in-person participants collaborated with a robot arm to build a tower (see \fig{front}). This experiment was motivated by our larger application of industrial settings where human workers must collaborate with robot partners. At each timestep the human picked up and added one block to the tower, $a_\mathcal{H}$, and then the robot stacked its block on top, $a_\mathcal{R}$. The state $s$ was the sequence of blocks in the tower. See videos of this user study here: \url{https://youtu.be/u8q1Z7WHUuI}}

\begin{figure}[t]
\vspace{0.5em}
	\begin{center}		
    \includegraphics[width=0.9\columnwidth]{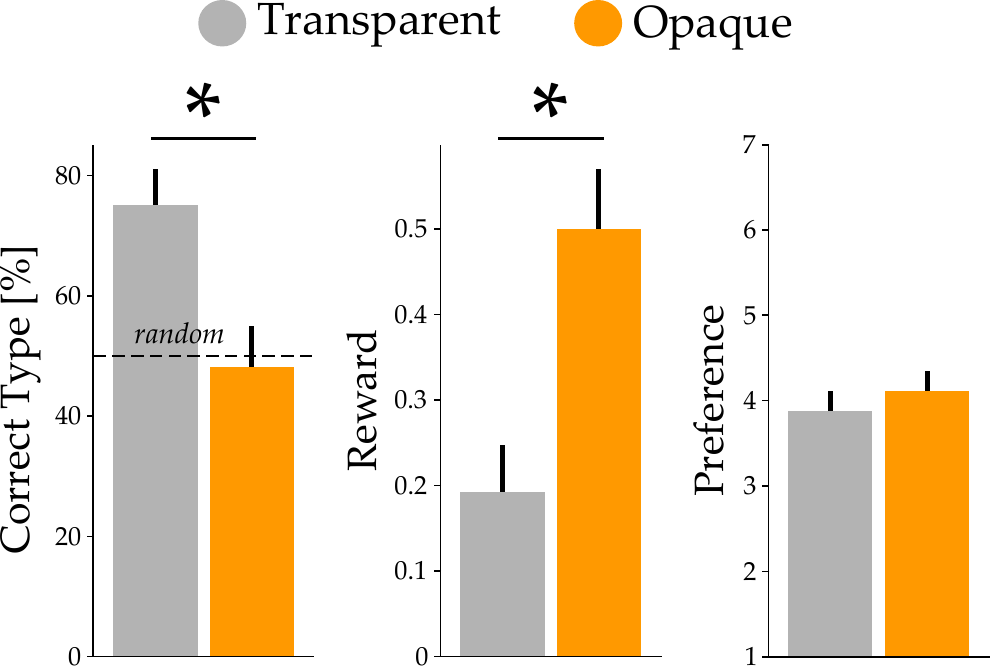}
		\caption{Results from our in-person user study. Participants collaborated with a robot arm to stack blocks (see \fig{front}). \textit{Correct Type} is the percentage of trials where users correctly identified that the robot was capable or confused: by guessing randomly users would be right $50\%$ of the time. As expected, the \textbf{Transparent} algorithm more effectively conveys the robot's type $\theta$ to the participants. However, when performing the collaborative assembly task with a physical robot arm, users achieved higher rewards with the \textbf{Opaque} algorithm. Users did not indicate a clear preference for either \textbf{Opaque} or \textbf{Transparent} partners. Error bars show standard error and an $*$ denotes statistical significance ($p<.05$)
    }
		\label{fig:block}
	\end{center}
\end{figure}

\p{Experimental Setup} We placed blocks of two different sizes and four different colors near the human and robot (see \fig{front}). The capable robot could pick up any of the blocks, while the confused robot was only able to stack the smaller blocks. At each timestep the human-robot team obtained a reward of $+0.5$ if they picked the same block, and $-0.5$ if they picked different blocks. Each taller block that was stacked on the tower added a reward of $+0.1$. In \fig{block} we normalize the interaction reward between $0$ and $1$ for ease of comparison with online user study results. To prevent the human from cheating (i.e., changing their block in response to the robot) the robot waited to move until after the human. To familiarize participants with the robot and alleviate any hesitation in interacting with it, we asked the participants to engage in an unrecorded practice run which did not include any algorithms from the study.

Each participant built towers with confused and capable \textbf{Opaque} robots, and with confused and capable \textbf{Transparent} robots. In total participants built $8$ towers where each tower contained $6$ blocks. We displayed the user's current score in real-time on a monitor next to the towers. 

\p{Participants and Procedure}
We recruited $13$ participants from the Virginia Tech community ($5$ female, ages $24.31\pm 3.45$ years). These participants provided informed consent under IRB$\#20$-$755$ and did not take part in the online study. We used a performance-based compensation model. Every user received a $\$10$ gift card for taking part in the study; for each tower they built with a reward higher than $10$ points, they received a performance bonus of $\mbox{\textcent}50$ (USD). We informed the participation about the performance based compensation before the study began. This compensation was designed to motivate participants to try and maximize their reward while working with the robot (i.e., to assemble the best tower possible).

After completing each tower we asked users if they could determine which type of robot they had worked with (i.e., if they could determine whether $\theta$ was capable or confused). Prior to the experiment, participants were told that in half of the interactions they would work with the confused robot, and in the other half of the interactions they would work with the capable robot. The capable and confused robots were also defined for the participants: users were told that the capable robot could pick up any block, but the confused robot could only pick up the smaller blocks. We grouped the towers into pairs of two, where each pair contained one \textbf{Opaque} robot and one \textbf{Transparent} robot. After every pair we asked if the user preferred the first robot or the second robot. We emphasize that users were never told which algorithm they were working with, or what the current $\theta$ was. The algorithms and robot types were presented in a randomized and counterbalanced order. Participants were never told which robots were confused and which robots were capable.

\p{Results} See our \href{https://youtu.be/u8q1Z7WHUuI}{video} and \fig{block}. To confirm that the \textbf{Opaque} algorithm withheld the robot's type --- and the \textbf{Transparent} robot revealed its type --- we first recorded the percentage of trials where participants correctly inferred whether the robot was confused or capable. For reference, users that guessed the robot's type $\theta$ completely at random would be right $50\%$ of the time. Our results in \fig{block} show that with \textbf{Opaque} user were not able able to determine robot's type: their responses were on par with this random baseline ($48\%$ correct). As expected, with the \textbf{Transparent} partner users identified $\theta$ correctly $75\%$ of the time. This supports our experimental design and suggests that the \textbf{Opaque} robot did indeed withhold the robot's type $\theta$, while the \textbf{Transparent} robot communicated this type to the participants.

We next proceeded to evaluate hypotheses \textbf{H1} and \textbf{H2}. Our results from \fig{block} are in line with the online user study: users reached higher rewards when working with \textbf{Opaque} partners ($p< .001$), and participants rated \textbf{Opaque} robots about the same as \textbf{Transparent} partners ($t(51)=.48$, $p=.63$). In practice, when working with the \textbf{Opaque} the robot would select the same block regardless of the robot's type $\theta$. This consistency may have helped participants predict how the robot would behave and enabled users to coordinate with the robot. By contrast, with \textbf{Transparent} the robot selected different blocks to add to the tower for each type $\theta$. By changing the block the robot communicated its type to the human. But this change may have also made it more difficult to anticipate what the robot would do, resulting in an increased number of interactions where the human was unsure how best to coordinate with the robot partner.

For hypothesis \textbf{H2}, we again found that the users did not have a clear preference for either \textbf{Transparent} or \textbf{Opaque} robots. We suggest that two factors may be at play here: (a) participants may prefer the \textbf{Opaque} robot because it leads to higher task reward, while simultaneously (b) participants may prefer the \textbf{Transparent} robot because it conveys its capabilities to the human. Perhaps this trade-off between performance and communication canceled out, and users were left with similar preferences for both algorithms. Regardless, we emphasize that our null effect for this experiment is not conclusive. We have not demonstrated that humans prefer \textbf{Transparent} and \textbf{Opaque} algorithms equally: we do not have enough evidence to claim that users prefer one of these options over the other.

\p{Discussion}
Our results across $13$ in-person participants suggest that robot arms in assembly settings can achieve higher rewards when withholding latent information. 
To support this finding, we checked whether \textit{learning effects} may have skewed our results. Because participants worked with the robot a total of $8$ times, it is possible that users picked up on patterns over the course of the experiment and then anticipated how the robot would behave when assembling their final towers.
Using paired t-tests we compared the final reward from the first four interactions and the last four interactions.
We found that the differences were not statistically significant with the \textbf{Opaque} robot ($t=0.622$, $p=0.54$) or the \textbf{Transparent} robot ($t=-1.33$, $p=0.20$). Hence, we conclude that there was not a significant learning effect during the experiment, and the towers that users assembled where not affected by repeated interactions.

For hypothesis \textbf{H2}, we again found that the users did not have a clear preference for either \textbf{Transparent} or \textbf{Opaque} robots. 
We suggest that two factors may be at play here:
(a) participants may prefer the \textbf{Opaque} robot because it leads to higher task reward, while simultaneously (b) participants may prefer the \textbf{Transparent} robot because it conveys its capabilities to the human.
Perhaps this trade-off between performance and communication cancelled out, and users were left with similar preferences for both algorithms.

Another possible factor which may have affected our results is how participants interpreted the robot’s actions. For example, because the \textbf{Transparent} robot changed which block it selected based on its current type, it is possible that participants found this robot to be antagonistic. We attempted to mitigate this factor by providing a clear explanation of the robot’s intent to the participants. Specifically, in our written instructions we informed participants that the robot is always collaborative, and “the robot is trying its best to help you.” We also explained that the capable robot can grasp any block, while the confused robot has limitations. Despite these efforts, users may have been affected by the perceived cooperation of the robots. This could have biased the participants to subjectively favor the \textbf{Opaque} robot, since its consistent behavior may have been seen as more collaborative. Overall, however, the participants’ preferences for \textbf{Opaque} and \textbf{Transparent} were not significantly different (see \fig{block}).


\section{Conclusion}

We considered settings where a robot is collaborating with a human, and both the human and robot share the same objective. 
Although prior work suggests that it is often optimal for collaborative robots to be transparent --- and take actions that convey their latent state --- we have introduced theoretical and experimental analysis to demonstrate that there are situations where transparency is not optimal.
We built on related works to formalize opaque robot behavior, and then developed a modified version of Harsanyi-Bellman \textit{ad hoc} coordination to identify optimal robot policies in stochastic Bayesian games. 
We proved that this optimal policy can be either fully or rationally opaque.
Our simulations suggest that opaque robot behavior is more likely to be optimal in settings where the human and robot collaborate across a short time horizon, or where the human learns gradually from the robot's actions.
We find experimental support for our analysis across two user studies with $43$ total users.
Within these experiments participants reached higher rewards when collaborating with opaque robots, and users did not perceive opaque robots as subjectively worse than their transparent counterparts.

\subsection{Limitations and Future Work}

One limitation of our experiments is that we focused on settings where there were two possible latent states (i.e., $N=2$). We designed our experiments with this binary choice in mind so that we could clearly explain the types to our participants: psychological research suggests that too much variability here can lead to random user choices \cite{erev2005adaptation}. 
Similarly, prior work on using transparent robot motion often communicates the robot's goal in settings where there are only two different options \cite{dragan2013legibility, dragan2015effects, bodden2018flexible}.
By restricting our user studies to $N=2$, we were able to mitigate user confusion and ensure that the experiments accurately tested the transparent and opaque algorithms.
Moving forward we are interested in conducting experiments with an increasing number of latent states.
We anticipate that --- during these experiments --- we may need additional measures of the human's cognitive workload and attention to understand whether opaque robot's simplify the human's decision making.

\subsection{Discussion and Applications}

There is a general trend in human-robot interaction to program robots so that their behavior is transparent; i.e., the robot should take actions that make its goal or intent clear to the human (see Section \ref{sec:related}, “Transparent Robots”). A variety of studies support this approach, and we agree that transparent robot behavior is often beneficial for human-robot interaction. However, the purpose of this manuscript was to explore the edge cases — is it always optimal or preferable for robots to select transparent behaviors? Our theoretical and experimental results provide a two part answer to this question. If the human agent is optimal, then transparent robot behavior will result in higher rewards when the human and robot interact for an extended period of time. But if the human does not know how to interpret the robot’s actions, or if the human and robot only interact briefly, then transparent robot behavior is actually \textit{inefficient}. Put intuitively, in these cases the robot is “wasting its time” trying to communicate with the human, and the robot should choose actions that help it efficiently complete the task, regardless of whether these actions are transparent or not. Our mathematical analysis enables designers to determine the crossover point where transparent behaviors become optimal.

Our findings should be applied to contexts where a robot and human are collaborating, and the robot can use its actions to convey its intent. This includes collaborative manufacturing settings (see the in-person user study in Section \ref{sec:user2}) as well as shared control settings (see the online user study in Section \ref{sec:user1}). Our results are particularly important for robots that will only interact with humans for a short period of time, because — in these shorter interactions — we theoretically and experimentally show that opaque behavior lead to better team performance. Here we see applications with autonomous cars (that will only be near a given human-driven vehicle for a short period of time) and with robots in public spaces, such as malls or stores (where members of the public may briefly interact with the robot).

Our work also applies to scenarios where the human is collaborating with a robot learner. For example, perhaps the robot learner has been trained offline, and now a human is going to deploy this robot within a cooperative task. The human user does not know a priori what exactly the robot learned: e.g., the robot might only have learned how to complete some subset of the task (such as attaching legs to a chair) as opposed to the entire task (assembling a chair). Here our game-theoretic analysis provides a way for the robot to select optimal actions while reasoning over the human’s uncertainty. If the human and robot are going to work together to assemble multiple chairs (i.e., have a long-term interaction), then the Harsanyi-Bellman ad hoc coordination method presented in Section 4 will likely control the robot to be transparent. Alternatively, if the human and robot are only going to assemble a single chair (i.e., have a short-term interaction), then our approach could control the robot learner to take opaque actions. In either case, our work provides the robot learner a way to assess whether it should take transparent actions to communicate its learned intent, goals, and capabilities, or whether the robot learner could help the human more efficiently by directly completing the task.



\backmatter



\section*{Declarations}

\begin{itemize}

\item Conflict of interest/Competing interests

No competing interests to declare.

\item Consent to participate
All participants were adults who were informed they could withdraw from the experiment at any time. They provided informed consent. Virginia Tech IRB$\#20-755$.

\item Availability of data and materials
Supplementary video is available at: \url{https://youtu.be/u8q1Z7WHUuI}

Simulation data is available at: \url{https://github.com/VT-Collab/opaque/}

\item Code availability 

Experimental and analysis codes are available in the following repository: \url{https://github.com/VT-Collab/opaque/}

\end{itemize}

\begin{appendices}

\section{}\label{secA1}

\begin{figure*}[t]
\vspace{0.5em}
	\begin{center}
    \includegraphics[width=1.5\columnwidth]{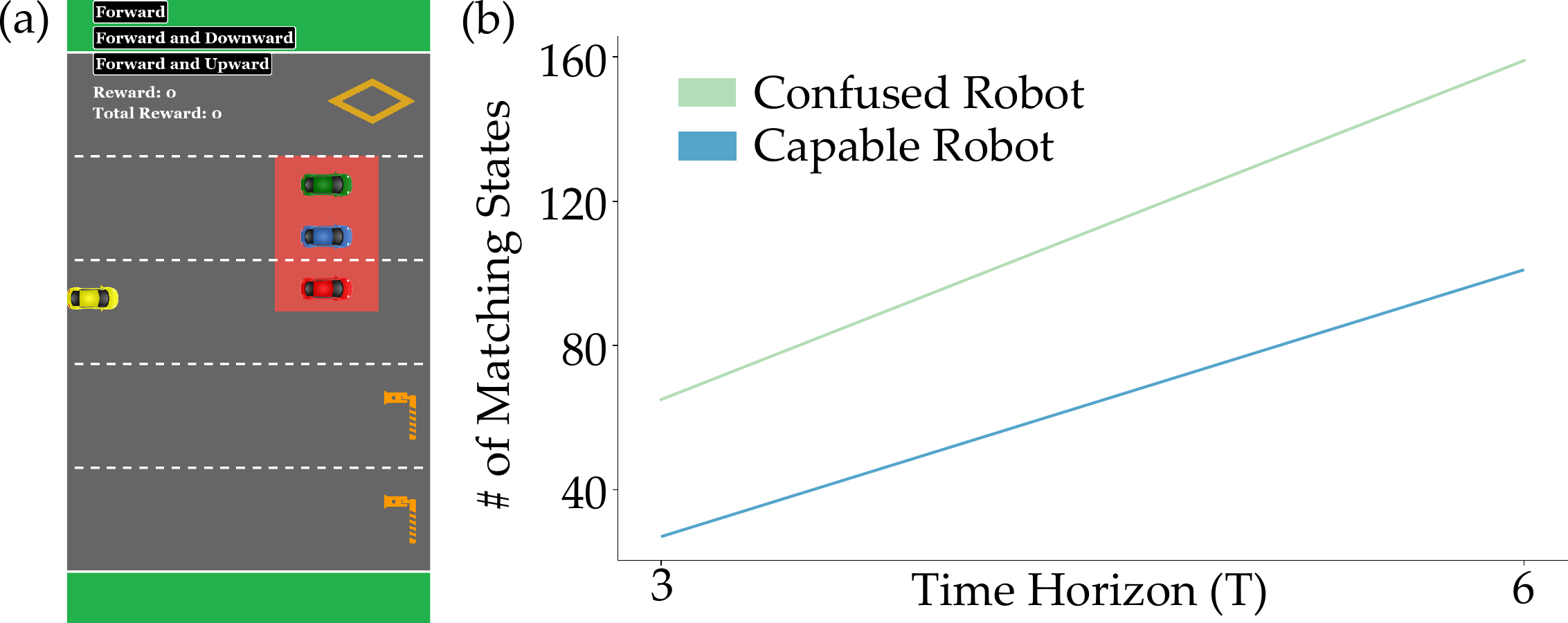}
		\caption{Results from our extended online user study. (a) Participants shared control with an autonomous agent to drive their vehicle along the highway and avoid other cars. Here the human’s car is shown in yellow, and the vehicles to avoid are surrounded by a red box. The human-robot team received rewards for making lane progress, staying on the road, and avoiding collision. (b) The $x$-axis captures the length of the interaction (in timesteps). The $y$-axis reports the number of times that the robot takes the same actions for the Opaque algorithm (our approach) and the Transparent algorithm (baseline).
    }
		\label{fig:A1}
	\end{center}
\end{figure*}

Our theoretical analysis in Section \ref{sec:method} suggests that it is optimal for robots to take opaque actions when they are only interacting with humans for a brief period of time. Our simulated results in Section \ref{sec:sims} support this analysis. When we apply our modified Harsanyi-Bellman ad hoc Coordination algorithm to simulated shared autonomy tasks, we find that opaque actions are more likely as the interaction time decreases (see \fig{sim1d} and \ref{fig:sim2d}). However, as the length of the interaction increases, the optimal behavior identified by our algorithm matches the policy of a transparent robot. Put another way, our approach theoretically leads to the same actions as a transparent robot during long-term interactions.

To move beyond these simulated results and test our theory with actual human users, we conducted a follow-up study. This follow-up study is an extension of the online user study reported in Section \ref{sec:user1}. Specifically, we focused on the Passing environment. Within this environment, participants shared control with an autonomous agent to drive their vehicle along the highway and avoid other cars (see \fig{A1}, (a)). Here the human’s car is shown in yellow, and the vehicles to avoid are surrounded by a red box. The human-robot team received rewards for making lane progress, staying on the road, and avoiding a collision.

The overall purpose of this follow-up study is to test whether our algorithm converges towards the transparent baseline as the length of the interaction increases. In Section \ref{sec:user1} of the manuscript we report our original results when the human and robot only interact for three timesteps (Time Horizon $T = 3$). Here we perform the same experiment with three timesteps and six timesteps (Time Horizon $T = 6$). We recruited 10 new participants using university mailing lists at Virginia Tech, the University of California, Berkeley, and the University of Illinois Urbana-Champaign. Each participant performed a total of $24$ passing interactions: these included working with two different algorithms (\textbf{Opaque} or \textbf{Transparent}), two different robot types (Confused or Capable), and two time horizons (three timesteps or six timesteps). The order of presentation was randomized and counterbalanced so that half of the users started each scenario with the \textbf{Opaque} robot and the other half started with \textbf{Transparent}.

Our results are summarized in \fig{A1}, (b). Here the $x$-axis captures the length of the interaction (in timesteps), and the $y$-axis reports the number of times that the robot takes the same actions for the Opaque algorithm (our approach) and the Transparent algorithm (the baseline). Aligned with our theoretical analysis, we found that the number of matching states increases as the time horizon increases. This supports our overall finding that optimal robot behavior converges towards transparent behaviors during longer interactions, but for brief interactions the robot’s optimal behavior may be opaque.

\end{appendices}

\bibliographystyle{plain} 
\bibliography{R1main-bibliography}

\end{document}